\documentclass[conference]{IEEEtran}
\IEEEoverridecommandlockouts
\usepackage{cite}
\usepackage{amsmath,amssymb,amsfonts}
\usepackage{graphicx}
\usepackage{textcomp}
\usepackage{xcolor}
\usepackage{comment}
\usepackage{url}
\usepackage{algorithm}
\usepackage{algpseudocode}

\def\BibTeX{{\rm B\kern-.05em{\sc i\kern-.025em b}\kern-.08em
    T\kern-.1667em\lower.7ex\hbox{E}\kern-.125emX}}

\begin{document}

\title{Integrating Causality with Neurochaos Learning: Proposed Approach and Research Agenda}


\author{\IEEEauthorblockN{Nanjangud C.~Narendra}
\IEEEauthorblockA{Artpark, Indian Institute of Science\\
Bangalore, India\\
ncnaren@gmail.com}
\IEEEauthorblockN{Nithin Nagaraj}
\IEEEauthorblockA{Complex Systems Programme \\National Institute of Advanced Studies (NIAS) \\
IISc. Campus, 
Bangalore 560 012, India\\nithin@nias.res.in}
}

\maketitle

\begin{abstract}

Deep learning implemented via neural networks, has revolutionized machine learning by providing methods for complex tasks such as object detection/classification and prediction. However, architectures based on deep neural networks have started to yield diminishing returns, primarily due to their statistical nature and inability to capture causal structure in the training data. Another issue with deep learning is its high energy consumption, which is not that desirable from a sustainability perspective.

Therefore, alternative approaches are being considered to address these issues, both of which are inspired by the functioning of the human brain. One approach is \emph{causal learning}, which takes into account causality among the items in the dataset on which the neural network is trained. It is expected that this will help minimize the spurious correlations that are prevalent in the learned representations of deep neural networks. The other approach is \emph{Neurochaos Learning}, a recent development, which draws its inspiration from the nonlinear chaotic firing intrinsic to
neurons in  biological neural networks (brain/central nervous system). Both approaches have shown improved results over just deep learning alone. 

To that end, in this position paper, we investigate how causal and neurochaos learning approaches can be integrated together to produce better results, especially in domains that contain linked data. We propose an approach for this integration to enhance classification, prediction and reinforcement learning. We also propose a set of research questions that need to be investigated in order to make this integration a reality.

\end{abstract}


\section{Introduction}\label{sec:intro}

Deep learning~\cite{lecun2015deep} comprises computational models that are composed of multiple processing layers to learn representations of data with multiple levels of abstraction. These layers are combined together to form what are called \emph{artificial neural networks (ANNs)} or \emph{neural networks (NNs)} for short. Deep learning has proved to be remarkably successful in complex tasks such as object detection, speech recognition, classification and prediction, and has been applied to various domains such as medical informatics, robotics, computer vision, etc.~\cite{dong2021survey}

However, it has become apparent that deep learning has started to yield diminishing returns~\cite{marcus2022deep}. There are a few reasons for this. First, deep learning relies on statistically derived correlations to produce its results, many of which could be spurious. Second, neural networks trained on a particular dataset are found to perform poorly on datasets from a different distribution. Third, since they comprise multiple processing layers, and need to be trained on large datasets, their energy consumption has already raised concerns~\cite{energy}. 

In order to address the issues of accuracy and energy consumption, alternative approaches are being considered. Prominent among them are \emph{causal machine learning} and \emph{Neurochaos Learning}. Causality~\cite{cavique2023causality} is concerned with the possible factors that could affect any object/event. For example, a pavement could have become wet due to either rain or a sprinkler. An example of an event would be -- the stock of XYZ went up because of a policy that was implemented in country ABC. Causality is increasingly being integrated into traditional deep learning methods to enhance their accuracy~\cite{berrevoets2023causal,luo2020causal}. This is because of the realization that humans think causally~\cite{vallverdu2024humans}. Neurochaos Learning (NL)~\cite{harikrishnan2021noise,sethi2023neurochaos} is based on the nonlinear chaotic firing exhibited by neurons in the brain. NL has been shown to be effective at classification tasks, even with a low number of training samples (sometimes with just a single sample/class for training). Therefore both causality and NL are brain-inspired. Indeed, in an earlier work~\cite{nb2022causality}, it has been shown that NL preserves the inherent causal structures present in input timeseries data. 

To that end, in this position paper, we investigate how these two newly discovered approaches - causality and NL - can be integrated together so that fullest advantage can be taken of both. Our particular emphasis is on domains containing \emph{linked data}~\cite{bizer2008linked}, i.e., data which is represented in the form of a graph. Prominent examples of such domains are Internet of Things (IoT), manufacturing, healthcare and life sciences, media and publishing and telecommunications. It is to be noted that the brain typically perceives data better when it is in a graph-like form, given the visual appeal of graphs.

We therefore view our position paper as one of the initial steps toward enriching the burgeoning area of \emph{neuromorphic computing}~\cite{schuman2022opportunities} that is being investigated as enhancing the field of Artificial Intelligence (AI) via brain-inspired techniques.

Our paper is organized as follows. In the next section, we present the background material that will be used in the rest of our paper. Our main contributions are presented in Section~\ref{sec:integration}, which comprise a mindmap of causality-NL integration, characterization of the causal model, causality modeling with graph neural networks as per the mindmap, and our proposed strawman approach to integrate NL into graph neural networks. Next, in Section~\ref{sec:extensions}, we will present and discuss two key extensions of our ideas, to predictive modeling and reinforcement learning. After that, as a postscript, we will present and discuss in Section~\ref{sec:onto} how integrating ontology modeling with causal models will enrich causal models, providing a further boost to causality-NL integration. Throughout our paper, in each section starting from Section~\ref{sec:integration} onward, we will also be presenting and discussing the key research questions that we see are pertinent to the subject matter of the section. Finally, our paper concludes in Section~\ref{sec:conclusions} with suggestions for future work. Throughout our paper, we will be illustrating our ideas using realistic illustrative examples.

\section{Background}\label{sec:background}

\subsection{Causality}\label{subsec:causality}

Causality is the principle that determines the factors and mechanisms that relates how one event or object influences  another event or object~\cite{pearl2018book}. It is derived from the intuition that humans often think in terms of ``cause and effect'' and how we perceive as the world operates under such a causal model. In~\cite{pearl2016causal}. Causality has been defined via the ``ladder of causation'' which has the following levels: association, intervention and counterfactual. Association defines the effect of $v_i$ on $v_j$, thus: $v_i \rightarrow v_j$, implying that $v_i$ is a cause of $v_j$. $v_i$ is typically referred to as an \emph{endogenous} variable. 

Sometimes other variables, also called \emph{exogenous} variables or \emph{confounding} variables, are also present, which could affect the causal relationship $v_i \rightarrow v_j$. Such variables need to be accounted for in order to accurately model $v_i \rightarrow v_j$, and this is done by fixing the value of the confounding variable, and then evaluating $v_i \rightarrow v_j$. This refers to the second rung of the causation ladder, i.e., intervention. Whereas, the third rung of the causation ladder, i.e., counterfactual, is about ``looking back'' and analyzing what $v_i \rightarrow v_j$ would have been had one or more confounding variables been intervened on, i.e., a type of ``what if'' reasoning.

Causal relationships for a set of variables can be combined together into a \emph{structural causal model (SCM)}~\cite{pearl2010introduction} that is essentially a directed acyclic graph with variables such as $v_i$ and $v_j$ as nodes and edges modeled via the causal relationship $v_i \rightarrow v_j$. Hence a SCM is a graph $G = (V, E)$, where $V$ is the set of variables, including exogenous variables, and each edge $e \in E$ is of the form $v_i \rightarrow v_j$. Edges can also be attributed via a probability value $p_{ij}, 0 < p_{ij} \le 1$, which refers to the relative strength of the causal relationship $v_i \rightarrow v_j$. The $p_{ij}$ value can be fixed or drawn from a probability distribution, enhancing the SCM into a Bayesian network~\cite{heckerman1998tutorial}.

A related concept to SCM is that of \emph{knowledge graph (KG)}, which is a representation of real-world entities and their relationships. A KG can therefore serve as an ontology of the domain in question. A KG does not model causal relationships, but can be augmented with causal relationships to form a \emph{causal knowledge graph (causal KG)}~\cite{jaimini2022causalkg}. 

\subsection{Neurochaos Learning}\label{subsec:nl}

Neurochaos Learning (NL)~\cite{harikrishnan2021noise} arose out of the observation that chaos and noise are inherent in the brain. Hence, like causality, NL is also a brain-inspired idea. The brain consists of $\approx 86$ billion neurons that interact with each other to form a complex network~\cite{azevedo2009equal}. These neurons are known to possess fluctuating responses to stimuli. This is partly due to their inherently chaotic nature and also due to noise, which is usually referred to in the literature as ``stochastic resonance (SR)''~\cite{benzi1982stochastic}.

In our earlier work, therefore~\cite{harikrishnan2020neurochaos,harikrishnan2021noise,balakrishnan2019chaosnet}, we have shown how SR can be used in NL. Also, the brain is known to be energy efficient and robust to noise, unlike computers. 

Our architecture, \textsf{ChaosNet}, consists of a
layer of chaotic neurons which is the 1D GLS (Generalized Lüroth
Series) map, $C_{GLS}: [0, 1) \rightarrow [0, 1)$, is defined as follows:\\

$C_{GLS}(x) =
\left\{
	\begin{array}{ll}
		\frac{x}{b}  & \mbox{if } 0 \le x < b \\
        \frac{1-x}{1-b} & \mbox{if } b \le x < 1
	\end{array}\label{eq:gls}
\right.
$\\

where $x \in [0, 1)$ and $0 < b < 1$.

Fig.~2 of Ref.~\cite{balakrishnan2019chaosnet} depicts the ChaosNet (NL) architecture. Each GLS neuron starts firing (from an initial neural activity of $q$ units) upon encountering a stimulus (normalized input data sample) and halts when the neural firing matches the stimulus (the topological property of the GLS maps guarantees this for almost every initial $q$). From the neural trace of each GLS neuron, we extract the
following features: firing time, firing rate, energy and entropy. These parameters are then fed into a classifier such as support vector machine (SVM)~\cite{noble2006support} (any other machine learning classifier, or even a neural network, can also be used). 

\begin{itemize}
    \item \emph{Firing time}: Time taken for the chaotic trajectory to recognize the stimulus
    \item \emph{Firing rate}: Fraction of time the chaotic trajectory is above the discrimination threshold $b$ so as to recognize the stimulus
    \item \emph{Energy}: For the chaotic neural trace (trajectory) $x(t)$ with firing time $n$, energy is defined as:

    \begin{equation}
        E = \sum_{t=1}^{n}|x(t)|^2.
    \end{equation}
    \item \emph{Entropy}: For the chaotic neural trace (trajectory) $x(t)$, we first compute the binary symbolic sequence $S(t)$ as follows:

        \begin{equation}
            S(t_i) =
            \left\{
            	\begin{array}{ll}
            		0  & x(t_i) < b, \\
                    1 & b \le x(t_i) < 1.
            	\end{array}
            \right.
        \end{equation}
        
        where $i = 1~to~n$ (firing time). We then compute Shannon Entropy of $S(t)$ as follows:

        \begin{equation}
            H(S) = - \sum_{i=1}^{2} p_i log_2(p_i) bits,
        \end{equation}

        where $p_1$ and $p_2$ refer to the probabilities of the symbols 0 and 1 respectively.
\end{itemize}


GLS neurons obey the topological transitivity property of chaos~\cite{balakrishnan2019chaosnet} and stochastic resonance~\cite{harikrishnan2020neurochaos} that enables them to perform classification tasks. We have demonstrated in an earlier work~\cite{nb2022causality} that NL preserves Granger causality of timeseries data, unlike Deep Neural Network architectures (LSTM, CNN) which destroy causal structure in the input training data during learning. This makes NL a very desirable candidate for integrating causal learning.


This further strengthens our intuition that NL and causality can be integrated to produce better results, especially when working with linked data~\cite{bizer2008linked}, i.e., data that is typically represented in a graph-like form. In what follows we will expand on this theme and present our ideas on causality-NL integration, as well as the research questions that need to be solved to make this integration a reality.

In addition, our earlier work on NL has produced other interesting and useful results. Firstly, we have shown in~\cite{harikrishnan2021noise} that intermediate levels of noise provide the best classification performance. Secondly, as shown in~\cite{sethi2023neurochaos}, NL can be combined with other machine learning classifiers to significantly enhance the performance of the latter.

\section{Integrating Causality and NL}\label{sec:integration}

\subsection{Causality-NL Integration: Mindmap}\label{subsec:mindmap}

\begin{figure*}
    \centering
    \includegraphics[width=1.0\linewidth]{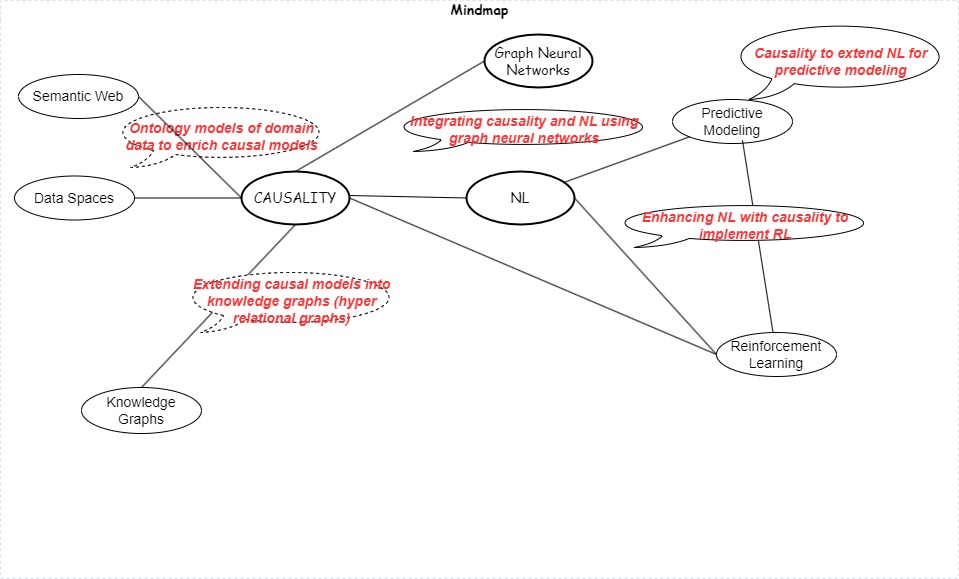}
    \caption{Causality-NL Integration Mindmap.}
    \label{fig:mindmap}
\end{figure*}

In this section, we present an overall mindmap (see Fig.~\ref{fig:mindmap}) of the various aspects of causality-NL integration.

The key aspects of our mindmap are as follows:
\begin{enumerate}
    \item The core of the mindmap is the integration between the ``CAUSALITY'' and ``NL'' nodes. We capitalize on the fact that neural networks are also graphs. In particular, since our focus is on linked data, we posit that \emph{graph neural networks (GNNs)}~\cite{zevcevic2021relating} would be most appropriate for working with linked data. Hence we build on preliminary work presented in~\cite{zevcevic2021relating} that relates SCMs and GNNs. In particular, the notion of intervention in a GNN is introduced in~\cite{zevcevic2021relating}. A useful result from~\cite{zevcevic2021relating} is that any GNN can be seen as a neural SCM variant, and~\cite{zevcevic2021relating} also presents an algorithm to construct a GNN based on an SCM.
    \item The other key aspects of the mindmap, are extensions of NL to predictive modeling~\cite{zhang2018link} and reinforcement learning~\cite{fathinezhad2023graph}, hitherto unexplored areas. 
    \item Another key aspect is the enrichment of the SCM itself using ideas from ontology engineering~\cite{kendall2019ontology}. It is expected that such enrichment would result in more accurate SCMs, thereby enhancing causality-NL integration. This has two parts: using semantic web and data spaces to enrich causal models~\cite{lassila2001semantic}; and using knowledge graphs to extend causal models into causal knowledge graphs as already mentioned above~\cite{jaimini2022causalkg}.
\end{enumerate}

\subsection{Causality Modeling with Graph Neural Networks}\label{subsc:gnns}

A detailed review of integration of causality and GNNs is presented in~\cite{job2023exploring}. GNNs that incorporate causality are referred to as Causal GNNS (CLGNNs). That paper identifies primarily two classes of CLGNNs: resolution-based and learning method-based. Within the resolution-based class, CLGNN-based methods have been developed for various types of input data: spectral, spatial, temporal and mixed. Within the learning method-based class, three types of learning methods have been addressed in the CLGNN literature:
\begin{itemize}
    \item Representation learning – building features that represent the structure
    \item Meta learning - solution for data scarcity by exploiting previously learned experiences towards learning an algorithm that generalizes across various tasks
    \item Adversarial learning - examining and devising defenses against adversarial attacks on models
    \item Reinforcement learning - causal mechanisms for agent learning process towards improved decision making
\end{itemize}

\subsubsection{Graph Neural Network Overview}\label{subsubsec:gnns}

But first we will describe what a GNN actually is. As described in~\cite{job2023exploring}, a GNN is a type of neural network that operates on data that is represented as a graph, i.e., linked data. It comprises two functions, $AGGREGATE$ and $COMBINE$, which are modeled as follows:

\begin{equation}\label{eq:gnn}
\begin{split}
    a_{v}^{k} = Aggregate^{k}\{H_{u}^{k-1}: u \in N(V)\}\\
    H_{v}^{k} = Combine^{k}(H_{v}^{k-1},a_{v}^{k})\\
\end{split}
\end{equation}

where the node representation is initialized as $H^{0} = X$, $N(v)$ is the set of neighbors of node $v$, K is the total number of GNN layers, $k = 1,2,...,K$ and $H^K$ is the finalized node representations. $a_{v}^{k}$ is the aggregated node feature of the
neighborhood $H_{u}^{k-1}$. In most GNN implementations, especially involving Graph Convolution Networks (GCNs), the $COMBINE$ function usually involves applying a non-linear activation function such as ReLU, sigmoid or tanh~\cite{broniatowski2003estimation}.

Graph learning happens at the following levels:
\begin{itemize}
    \item Node level: node properties are predicted. Node-level features may be importance-based or structure-based.
    \item Graph level: the features of the entire graph structure are captured for computing graph similarities.
    \item Link or Edge level: existing links are used to make predictions on new or missing links. Some link-level features are distance-based, local neighborhood overlap-based or global neighborhood overlap-based.
\end{itemize}

Graph learning tasks are of the following types:
\begin{itemize}
    \item Node Classification is the task of determining the class of a node based on classes of neighboring nodes.
    \item Graph Classification is the task of classifying an entire graph into various groups.
    \item Node Regression is the task of predicting a continuous value for a node.
    \item Link Prediction predicts potential relationships between two nodes in a graph.
\end{itemize}

Classification in GNNs follows the following process. First, a GNN layer computes the input features from the input graph and generates node embeddings by aggregating the features. Node features are updated with an update function and the transformed graph is passed through a classification layer to predict labels. The most used GNN types are Graph Convolution Networks (GCN), Graph Attention Networks (GAT), Graph Autoencoders (GAE) and Graph Isomorphism Networks (GIN). Further details of these GNN types are available in~\cite{job2023exploring}.

\begin{figure}
    \centering
    \includegraphics[width=0.5\linewidth]{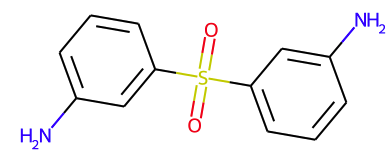}
    \caption{Example of a molecule - from~\cite{gnn-example}.}
    \label{fig:molecule}
\end{figure}

A well-known example of graph classification would be drug discovery, wherein a key task would be molecular property prediction~\cite{moleculenet}, in particular, whether a molecule is toxic or not. One such example molecule is depicted in Fig.~\ref{fig:molecule} and is derived from an online GNN tutorial~\cite{gnn-example}.
Some node features of the graph representation of the molecule,  could be: atomic number, degree, formal charge, hybridization, is aromatic (boolean), is in ring (boolean). Some edge features, that define bonds between atoms (nodes), could be type of bond, is conjugated (boolean), stereochemical configuration. Hence the typical approach to be followed, to classify the molecule as per its toxicity, would be: (a) embed nodes using message passing based on the graph's node and edge features; (b) aggregate these into a single graph embedding (using a model such as the one described in Equation~\ref{eq:gnn}, and (c) train a classifier based on the graph embedding. For this example, the classification would be to detect the toxicity of the molecule based on the node and edge features of its graph representation. If the graph representation in question is too small for a single GNN, then multiple batches of graphs can be integrated and fed to the GNN simultaneously, as shown in~\cite{graph-class}. Details of a sample implementation of this approach are presented in~\cite{gnn-example}.

Among graph representation learning methods, of particular interest for us would be C-GraphSAGE~\cite{zhang2022causal}. This is a graph classification method based on causal sampling. An enhancement of the well-known GraphSAGE inductive representation learning technique~\cite{hamilton2017inductive}, C-GraphSAGE improves upon GraphSAGE by using causal sampling instead of random sampling. Briefly, C-GraphSAGE improves the robustness of classification via what it calls ``causal bootstrapping'', a method that computes weights between the neighbors of a target node and their labels. Target node embedding is subsequently accomplished by aggregating over the network.

\begin{figure*}
    \centering
    \includegraphics[width=1.0\linewidth]{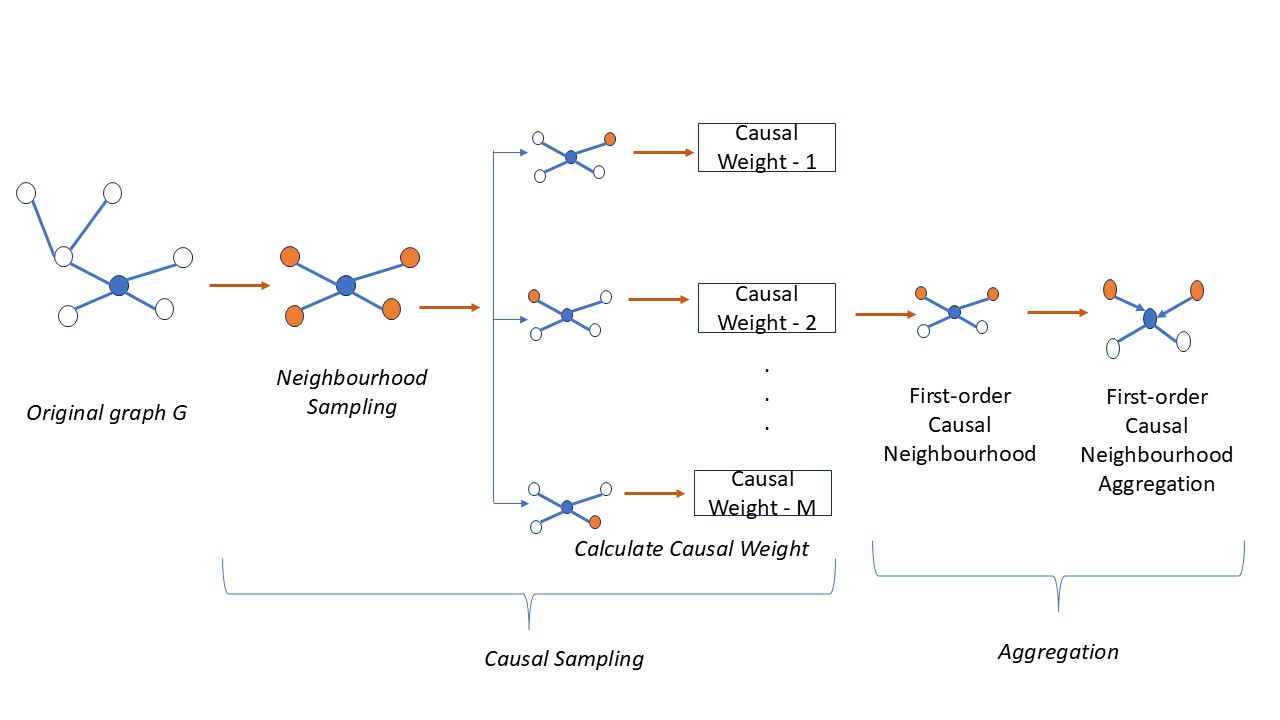}
    \caption{C-GraphSAGE process.}
    \label{fig:cgraphsage}
\end{figure*}

C-GraphSAGE is pictorially depicted in Fig.~\ref{fig:cgraphsage}. The process comprises two steps: causal sampling and aggregation. 
If we denote a graph as $G = (V, E)$, with $V$ being the set of nodes and $E$ being the set of edges, and $N(v)$ being the set of $v$'s neighbors, the key aspect of C-GraphSAGE over GraphSAGE~\cite{hamilton2017inductive}, is that it takes causal information into account between the nodes and their labels. 

Hence referring to Fig.~\ref{fig:cgraphsage}, after the initial neighborhood sampling, causal sampling is performed. This is done by considering the fact that for any node pair of nodes $v_i$ and $v_j$, if there is a path $<v_i, v_k, v_j>$, then the nodes in $N(v_k) \cup v$ could causally affect each other.

\begin{figure}
    \centering
    \includegraphics[width=1.0\linewidth]{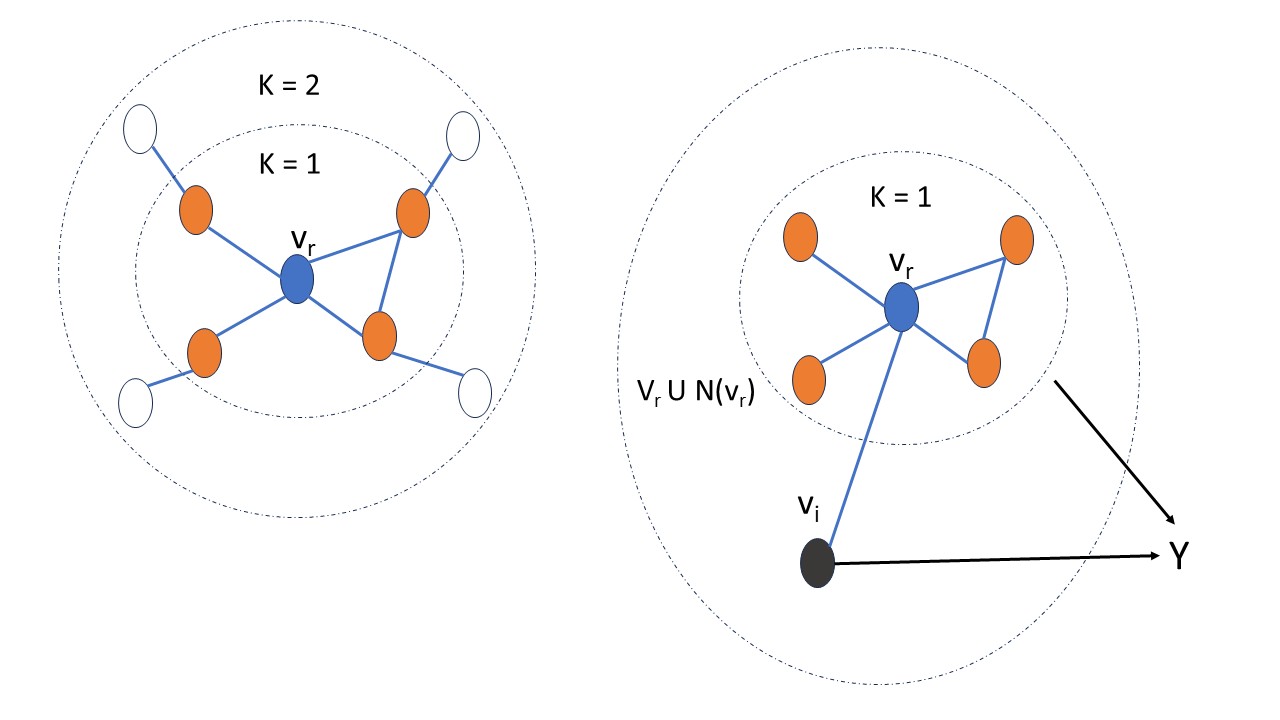}
    \caption{Neighborhoods Sampling and Mapping to the Label Set.}
    \label{fig:neighborhood}
\end{figure}

Hence in order to perform causal sampling, a suitable structural causal model, is needed which can be constructed from the neighbors of $v_k$ as depicted in Fig.~\ref{fig:neighborhood}. Assuming $y_i$ is the label of node $v_i$, which needs to be predicted, $v_i$ will be the cause of $y_i$ as shown in Fig.~\ref{fig:neighborhood}. Hence any node in the set that includes $v_k$, $v_k$'s neighbors, but excluding $v_i$, along with $v_i$, is the cause of $v_i$'s label in $Y$. Hence there are two causal paths: (i) $v_i \rightarrow Y$ and (ii) $v_k \cup N(v_k) \setminus v_i \rightarrow Y$.

So, as per~\cite{zhang2022causal}, causal information can be added to an undirected graph which did not represent causality earlier, in a manner consistent to the approach described in~\cite{hamilton2017inductive}.

Hence due to the existence of the above two causal paths $v_i \rightarrow y_i$ and $\overline{V_i}(v_r) \rightarrow v_i$ (see Fig.~\ref{fig:neighborhood}), the causal path $\overline{V_i}(v_r) \rightarrow y_i$ clearly exists. In this way, in~\cite{zhang2022causal}, the undirected data graph with no causal information is imbued with directed, causal information, in a manner consistent with the GraphSAGE~\cite{hamilton2017inductive} method on which it is built.

\subsection{Integrating Neurochaos Learning into Graph Neural Networks}\label{subsec:nlgnns}

Based on the above discussion, we envisage the following (non-exhaustive) list of possibilities for integrating NL into GNNs:
\begin{itemize}
    \item Replace the $COMBINE$ function of Equation~\ref{eq:gnn} with a neural spike such as 1D GLS map (Equation~\ref{eq:gls}) or Logistic Map~\cite{phatak1995logistic}. Another alternative that could be investigated, is the SR function presented in~\cite{manuylovich2024robust}, which is actually a stochastic ordinary differential equation. (Indeed,~\cite{manuylovich2024robust} has demonstrated robustness of neural networks with noisy datasets using its SR function.)
    \item A variant of the above, in a manner similar to that presented in~\cite{lansdell2023neural}, is to model the neural spikes so that the neurons can themselves estimate their causal effects. The neuron can then use this effect to estimate its gradients and calculate its synaptic strengths. As shown in~\cite{lansdell2023neural}, this approach enables the neural network to learn to maximize reward, especially in the presence of confounded inputs. 
    
    In a similar vein, spike-induced GNNs, as introduced in~\cite{chen2025signn}, can also be investigated. The approach in that paper uses a variant of the well-known leaky integrate-and-fire neuron~\cite{lif} as the activation function in order to implement graph representation learning.
    
    \item Directly introduce SR in the edge weights of the GNN, turning the graph of the GNN into a stochastic graph~\cite{rezvanian2016stochastic}. Neural spikes such as those introduced above can be used to introduce stochasticity in the weights. It is to be noted, however, that the method in~\cite{rezvanian2016stochastic} confines itself to calculating network measures in social networks. Hence investigating how it can be extended to other domains based on linked data, would be needed.
    \item Tailor the approach of SR introduction to the edge weights in a \emph{causal} manner. That is, considering that, as presented in~\cite{zevcevic2021relating}, a GNN can be represented as a variant of a causal graph, coordinate the introduction of SR into the weights so as to correspond to a causal path among the nodes of the graph. Intuitively, this is actually quite feasible, since a causal relationship between two nodes in a causal graph is already represented using Bayesian probability, which anyway models uncertainty. Moreover, this structure would also mimic how multiple neurons would interact with each other, via synapses. The key here would be determine the appropriate method for coordinated SR introduction. 
    
    The technique presented in~\cite{luo2020causal} can also be considered here, where the emphasis is on eliminating confounders in timeseries data via random sampling and recombining variant patterns to create an intervention distribution. The objective is to eliminate the effect of hidden confounders. The key idea behind~\cite{luo2020causal} is to use a spatio-temporal GNN, with the following modules: a causal feature learning module, a multi-layer spatiotemporal convolution module, a causal intervention module, and a predictive output module. Of special interest is the causal feature leaning module, which separates causal and non-causal features of the timeseries data.

    \item In this context, the Graph Neural Stochastic Differential Equations (GN-SDE) technique presented in~\cite{bergna2023graph} is worth noting. This enhances the traditional Graph Neural Ordinary Differential Equation (GN-ODE) technique by adding a diffusion function to the differential equation that models the stochasticity. This approach allows for the inclusion of prediction uncertainty, and~\cite{bergna2023graph} has shown via empirical studies that it can handle out-of-distribution datasets as well. This approach could be investigated as an alternative or supplement to the SR function approach presented in~\cite{manuylovich2024robust}. 

    \item Another approach to consider, could be built on the model proposed in~\cite{fang2021brain}. That model works with leaky integrate-and-fire neurons running in spiking neural network, and proposes that neuron firing be coordinated with the causal graph that represents the data. That is, given two nodes $A$ and $B$ in the causal graph, with the causal connection $A \rightarrow B$, each node represents a population of neurons. This causal connection is then transformed into whether the neuron population discharges, in the order defined by the cause-and-effect relationship defined in the causal graph. Extending this approach to GNNs could be investigated.
    
\end{itemize}

Based on the aforementioned possibilities, and the earlier introduction to the prior work, the following research questions arise:\\\\
\noindent \textbf{RQ1}: How would traditional graph embeddings, used in general GNNs so far, be modified to accommodate SR-based approaches such as NL?\\\\
\noindent \textbf{RQ2}: As already posed in~\cite{job2023exploring}, how can causality in dynamic environments be integrated with NL? In particular, capturing causal dynamics in a continuous environment is quite difficult, due to the difficulty in gathering timeseries data to account for the changes in the system. Also, if the causality has a spatial component, incorporating spatial causality also poses a challenge. Perhaps the method proposed in~\cite{fang2021brain} can be considered here.\\\\
\noindent \textbf{RQ3}: How would SR be introduced directly into GNNs? One approach could be our strawman approach introduced above. Alternate approaches that could be explored, involve introducing random features~\cite{sato2021random} or stochastic aggregation~\cite{wang2021stochastic}.\\\\
\noindent \textbf{RQ4}: Causal graphs typically come with Bayesian probabilities that define the possible cause of a variable on another. How could they be incorporated into our approach? In this regard, would probabilistic graphical models as described in~\cite{feldstein2024mapping} be useful? If so, how?\\\\
\noindent \textbf{RQ5}: Causal graphs can be quite large as already stated in~\cite{job2023exploring}, especially if augmented to become causal knowledge graphs~\cite{jaimini2022causalkg}. How can they be partitioned so that the overall problem can be solved via a ``divide and conquer'' approach? And if such an approach is indeed possible, how would the results be combined? Would it be perhaps via the approach suggested in~\cite{job2023exploring}, i.e., consolidate information from multiple graphs via representation learning?\\\\
\noindent \textbf{RQ6}: Alternatively, how to solve the \emph{reverse} problem of \textbf{RQ5}, i.e., how to use NL-based techniques to uncover these probabilities themselves? Could the fact that NL preserves causality~\cite{nb2022causality} be useful here, and if so, how?

\section{Extensions}\label{sec:extensions}

In this section, we discuss some extensions beyond the usual classification-based neural network-based approach. Our focus will be on predictive modeling and reinforcement learning.

\subsection{Extensions to Predictive Modeling}\label{subsec:pred}

As discussed in~\cite{zhang2018link}, there are plenty of heuristic solutions for link prediction for network-structured (i.e., linked) data. First-order heuristics typically involve one-hop neighbors of the target nodes~\cite{barabasi1999emergence}. Second-order heuristics~\cite{zhou2009predicting} are also sometimes considered. This can also be extended to higher-order heuristics that require knowledge of the entire network. 

However, the approach in~\cite{zhang2018link} adopts a different strategy. From the subgraphs of the graph representing the data, it extracts graph structure features using a neural network. This is subsequently used for link prediction.

\begin{figure}
    \centering
    \includegraphics[width=1.1\linewidth]{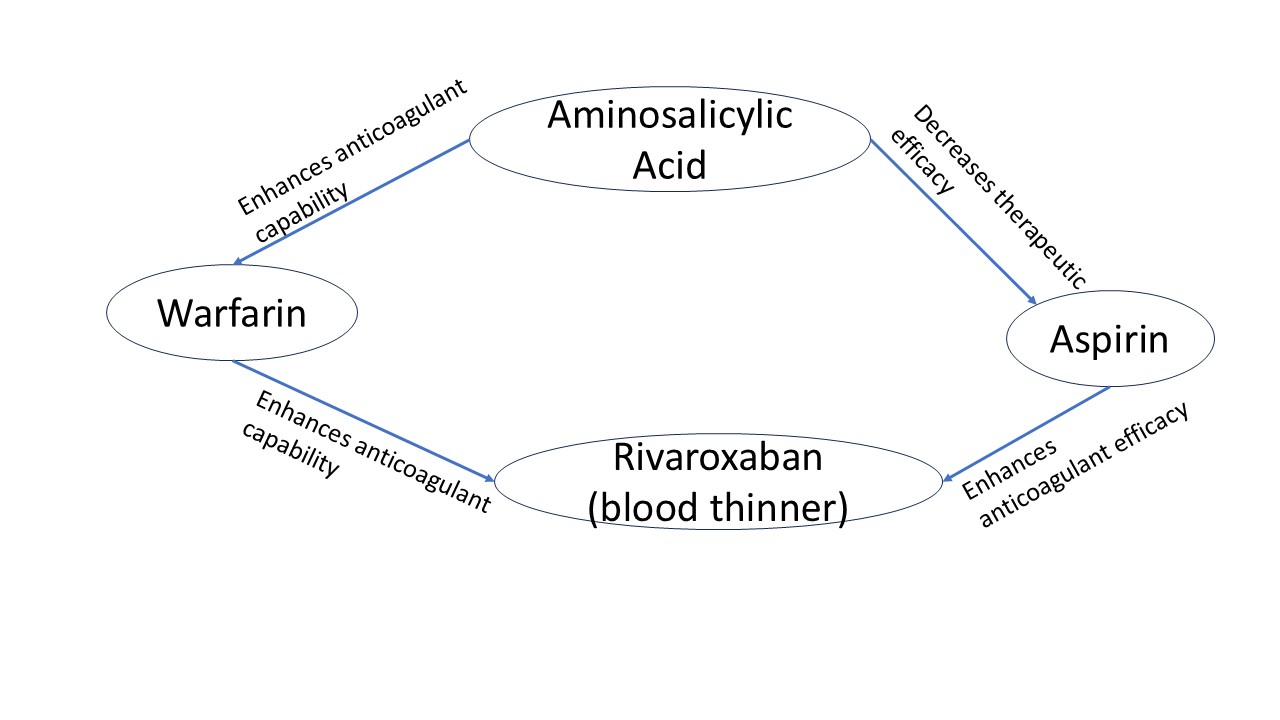}
    \caption{Drug-Drug Interaction example}
    \label{fig:drug}
\end{figure}

A well-known example of prediction related task would be to predict drug-drug interactions~\cite{du2024customized}, a common requirement in pharmacology. For such a task, drugs are represented as nodes and their interactions as edge in the graph, and the aim is to predict any missing edges, i.e., drug-drug interactions based on the existing data encoded in the graph. One such example subgraph from~\cite{du2024customized}, is depicted in Fig.~\ref{fig:drug}. In Fig.~\ref{fig:drug}, to predict the interaction between Aspirin and Warfarin, it can be seen that the therapeutic efficacy of Aminosalicylic Acid can decrease when combined with Aspirin, suggesting that Warfarin and Aspirin could be similar. Also, since Aminosalicylic acid may increase the anticoagulant capability of Warfarin and also since Aminosalicylic Acid resembles Aspirin, the method in~\cite{du2024customized} infers that Aspirin may also increase the anticoagulant capability of Warfarin. Therefore this predicts the link between Aspirin and Warfarin.

Based on the above discussion on link prediction, the following research questions arise:\\\\
\noindent \textbf{RQ7}: How would heuristics such as those described in~\cite{zhang2018link,du2024customized} be incorporated into an NL-based neural network to perform predictive modeling in graph-based data? Perhaps summation and activation functions would need to be remodeled to incorporate predictive modeling.\\\\
\noindent \textbf{RQ8}: Here too, similar to \textbf{RQ4}, how would Bayesian probabilities be incorporated here to provide predictions with a certain degree of confidence?

\subsection{Extensions to Reinforcement Learning}\label{subsec:rl}

Neural networks have been used in reinforcement learning (RL) for a long time~\cite{stanley2002efficient}.  The motivation for using neural networks for RL instead of the traditional approach using Q-tables, is that, for large data sizes, Q-table size could become large and unwieldy. In recent times, GNNs have also been used for reinforcement learning, as documented in~\cite{fathinezhad2023graph}. The typical process of using GNNs in RL, is as follows. First, the local observation of agents is encoded into the feature vector in the embedded layer. Second, a graph attention network is used to define the edge weights as the strength of the connection in the coordination graph between each agent and its neighbors. Thirs, graph convolution is applied to perform message across all agents. Finally, the deep Q-network is used to approximate the Q-function.

The next action for the agents is determined based on the maximum output of the Q-network.

As an illustrative example of RL, remaining in the pharmacological domain as per our earlier two examples, the graph-based method in~\cite{atance2022novo} uses RL in drug discovery. Building on a pre-trained generative model, this method uses RL to guide the model to create molecules that obey a particular profile, even if the molecules were not present in the training set. The pre-trained model is generated based on GraphINVENT~\cite{mercado2021graph}, which uses GNNs to generate molecular graphs.

Based on the above discussion on using GNNs for reinforcement learning, the following research questions arise:\\\\
\noindent \textbf{RQ9}: How can NL-enhanced GNNs be used to automatically generate Q-values, via SR? Perhaps the method presented in~\cite{mercado2021graph} could be used as a basis for this.

In addition, a method for reinforcement learning using spiking neurons, has been presented in~\cite{kiselev2024purely}; perhaps that could be enhanced by adapting it for NL-enhanced GNNs.\\\\
\noindent \textbf{RQ10}: How would the graph-based nature of the data in GNNs be represented here? 

\section{Ontology Modeling to Enrich Causal Models}\label{sec:onto}

Ontology modeling is the basis of building causal models, and helps represent causal models accurately. Hence an ontology model can be seen as a schema of a causal graph. Thus we believe that focus on ontology modeling is needed to develop accurate causal models that can in turn help improve causality-NL integration.

Integrating ontologies and causal models is an active area of research. In~\cite{jaimini2024causallp}, the authors present CausalLP, an approach that formulates the issue of incomplete causal networks as a knowledge graph completion problem. More specifically, the task of finding new causal relations in an incomplete causal network is mapped to the task of knowledge graph link prediction. The use of knowledge graphs to represent causal relations enables the integration of external domain knowledge; and as an added complexity, the causal relations have weights representing the strength of the causal association between entities in the knowledge graph. 

CausalLP has four primary phases: 1) encoding the causal associations in data as a causal network, 2) translating the
causal network into a causal knowledge graph, 3) learning knowledge graph embeddings from the causal knowledge graph,
and 4) using the knowledge graph embeddings for causal link prediction tasks.

On similar lines, an inductive link prediction approach using semantics is presented in~\cite{liu2024relation}. This approach, via a random walk strategy and a graph attention network, calculates the combined structural and semantic scores of neighbors of nodes. Through this, the approach forms a subgraph with key nodes, which integrates and represents information about node neighbors.

On a related but different note, a higher-order link prediction for hypergraphs is presented in~\cite{rui2024higher}. The reason for considering hypergraphs instead of just graphs, is that causal knowledge graphs would be hypergraphs, with multiple edge types between any pair of nodes. In~\cite{rui2024higher}, the solution to this problem is presented via hypergraph neural networks (HGNNs), which are variant of GNNs. Since typical HGNNs could get unwieldy, the approach in~\cite{rui2024higher} presents a model of a light HGNN (LHGNN), performs hybrid aggregation to obtain hyperlink embeddings at a level higher than link embeddings.

Based on the above discussion, the following research questions arise:\\\\
\noindent \textbf{RQ11}: Can NL-enhanced GNNs help uncover hidden links/hyperlinks faster?\\\\
\noindent \textbf{RQ12}: How can the causal knowledge graph be partitioned, in a manner posed in \textbf{RQ5}, to allow for faster hyperlink prediction using less memory and storage?\\\\
\noindent \textbf{RQ13}: How can these techniques be extended in the case of temporal knowledge graphs~\cite{cai2022temporal}, i.e., those that model streaming data such as in IoT or social media domains?\\\\
\noindent \textbf{RQ14}: How can these techniques be used to evolve knowledge graphs, so that they in turn can provide graph-based inputs to NL-enhanced GNNs to perform classification and prediction?

\section{Conclusions}\label{sec:conclusions}
In this position paper, we investigated two crucial machine learning techniques, viz., causal learning and neurochaos learning (NL), that can help enhance deep learning by addressing its key issues, viz., statistical nature, reliance on large datasets and high energy consumption. Motivated by our recent results that show that NL does preserve causality in timeseries data, we investigated how causality and NL can be integrated together for better results. By employing graph neural networks which can model linked data, we showed how causality and NL can be integrated together. And to demonstrate the richness of the topic of causality-NL integration, we presented and discussed several research questions that arise. We believe that solving these research questions will help make causality-NL integration a reality.

Our future work would be to investigate the posed research questions, and develop and demonstrate techniques that can help solve them. This would also involve further investigating and taking forward our proposed approach described in Section~\ref{subsec:nlgnns}.

\bibliographystyle{IEEEtran}
\bibliography{refs}

\end{document}